\documentclass[sigconf]{acmart}
\usepackage{amsmath}
\usepackage{comment}
\usepackage{newtxtext,newtxmath}
\usepackage{setspace}
\usepackage{nicematrix}
\usepackage{tikz}
\usepackage{booktabs}
\usepackage{multirow}
\usepackage{diagbox}
\usepackage{algorithm}
\usepackage{adjustbox}
\usepackage{algpseudocode}
\usepackage{graphicx}

\AtBeginDocument{%
  \providecommand\BibTeX{{%
    \normalfont B\kern-0.5em{\scshape i\kern-0.25em b}\kern-0.8em\TeX}}}

\copyrightyear{2023} 
\acmYear{2023} 
\setcopyright{acmlicensed}\acmConference[CIKM '23]{Proceedings of the 32nd ACM International Conference on Information and Knowledge Management}{October 21--25, 2023}{Birmingham, United Kingdom}
\acmBooktitle{Proceedings of the 32nd ACM International Conference on Information and Knowledge Management (CIKM '23), October 21--25, 2023, Birmingham, United Kingdom}
\acmPrice{15.00}
\acmDOI{10.1145/3583780.3615269}
\acmISBN{979-8-4007-0124-5/23/10}

\begin{document}

\title{Lightweight Adaptation of Neural Language Models via Subspace Embedding}

\author{Amit Kumar Jaiswal}
\authornote{Work done when the first author was at UCL.}
\orcid{0000-0001-8848-7041}
\affiliation{%
  \institution{University of Surrey}
  \country{United Kingdom}
}
\email{amitkumarj441@gmail.com, a.jaiswal@surrey.ac.uk}

\author{Haiming Liu}
\orcid{0000-0002-0390-3657}
\affiliation{%
  \institution{University of Southampton}
  \country{United Kingdom}}
\email{h.liu@soton.ac.uk}

\renewcommand{\shortauthors}{Amit Kumar Jaiswal \& Haiming Liu}

\begin{abstract}
Traditional neural word embeddings are usually dependent on a richer diversity of vocabulary. However, the language models recline to cover major vocabularies via the word embedding parameters, in particular, for multilingual language models that generally cover a significant part of their overall learning parameters. In this work, we present a new compact embedding structure to reduce the memory footprint of the pre-trained language models with a sacrifice of up to 4\% absolute accuracy. The embeddings vectors reconstruction follows a set of subspace embeddings and an assignment procedure via the contextual relationship among tokens from pre-trained language models. The subspace embedding structure\footnote{Code: https://github.com/amitkumarj441/CIKM2023\_SubspaceEmbedding} calibrates to masked language models, to evaluate our compact embedding structure on similarity and textual entailment tasks, sentence and paraphrase tasks. Our experimental evaluation shows that the subspace embeddings achieve compression rates beyond 99.8\% in comparison with the original embeddings for the language models on XNLI and GLUE benchmark suites.
\end{abstract}





\maketitle
\vspace{-10pt}
\section{Introduction}
\begin{figure}
\includegraphics[width=\linewidth]{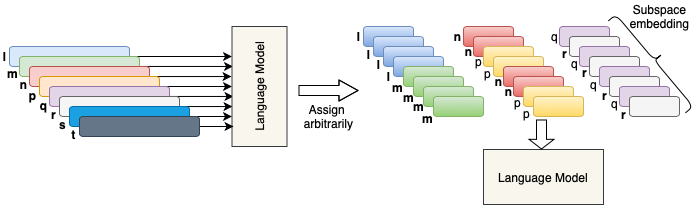}
\caption{Pictorial representation of subspace embedding. Given the language model with eight embedding vectors (leftmost) to be divided into three subspace embedding vectors. The naming convention of the subspace embedding blocks follows identical letters as they share learning parameters throughout the embeddings.} \label{fig:subspacee}
\vspace{-20pt}
\end{figure}

Representation of information contents by means of a geometrical relationship via embeddings plays a central role in neural networks, including language models, attention-based models, and graph neural networks. In neural language models (NLMs), each contextual component is catered to by a contextual embedding. Word2Vec embeddings~\cite{mikolov2013distributed} generate embedding vectors at the word level and so a paucity of out-of-vocabulary (OOV) problems arise from diversified words. Then, a popular language model using subword information~\cite{bojanowski2017enriching} has been developed to subdivide words into various segmented words. Enriching subwords spans different tokenizers~\cite{sennrich2016neural,kudo2018sentencepiece,tay2021charformer} that deal with words data to generate common tokens. In addition, each token comprises a varied context and possesses an inadequate criterion for distributing the tokens. For instance, tokens such as $\langle \text{O} \rangle$ and $\langle \text{o} \rangle$ prescribed to varied embeddings share important generic properties. These atomic-level tokens such as `Om' (in English) manifested as `Aum' in the Hindi language character, i.e. both contain vowels and common alphabetic characters with explicit identical meaning. Words depicting identical meanings have their tokens recline to interpolate in an embedding space, provided the language model is trained with such tokens. This paper proposes a new embedding structure where word embedding is decomposed as multiple subspace embeddings. Subspace embedding delineates the latent space of contextual elements within a token, and it can be defined for each element that composes to form the original embedding. Thus, the original embedding vector comprises subspace embeddings (i.e. shared between vocabularies) that play a part in employing the common learning parameters with closely situated embedding vectors. A subspace of an embedding~(\(E\)) is any subset of \(E\) that is also itself an embedding space, so-called the subspace embedding (it is different from the oblivious subspace embedding \cite{sohler2011subspace}, neural network subspace~\cite{wortsman2021learning}). Different sporadic subspace embeddings characterise based on their structural topology. Also, the neural network subspace is a learning and optimisation mechanism for deep networks. The subspace embeddings create an arbitrary-sized vector of each word that incorporates semantic relationships. In the initial process of embedding compression, we arbitrarily assign the subspace embedding to each token based on its index and perform a Cartesian product with subspace embedding to construct embedding vectors (shown in Fig.~\ref{fig:subspacee}). However, mapping these subspace embeddings does not signify the token's context. To overcome the mapping of subspace embeddings, we employ K-means clustering to distinguish non-overlapping tokens in the embedding space via RoBERTa~\cite{liu2019roberta}, a pre-trained language model.\\
\textbf{Main Findings:} Our proposed method is for word embeddings compression in pre-trained language models (PLMs). We found that our approach substantially alleviates the number of learning parameters in the embedding part with the usage of the Cartesian product. Also, applying subspace embedding solves the out-of-vocabulary problem in the language models. Additionally, our proposed approach can be used for PLMs by substituting the input embedding through subspace embedding. We conduct an extensive evaluation of our proposed compact embedding structure on English and multilingual datasets. Our main structure of the pre-trained language model for downstream tasks follows RoBERTa~\cite{liu2019roberta}. Also, we employ XLM-R~\cite{conneau2020unsupervised} for performance tests of subspace embedding on multilingual datasets.
\vspace{-10pt}
\section{Related Work}
\textbf{Word Embeddings:} In Word2vec, the vocabulary comprises words based on the input data which encounter the problem of OOV. However, certain language models~\cite{pennington2014glove,bojanowski2017enriching} segment the words into sub-tokens to learn words co-occurrence. Consecutively, attention-based models appeared to embed the semantics in longer sentences. This new generation of self-trained models is led by architectures such as ELMo~\cite{peters-etal-2018-deep}, which collect the embeddings of the bidirectional language models and hidden states to unleash contextual embeddings. In the case of the attention-based language models~\cite{kenton2019bert,liu2019roberta,lan2020albert} based on transformer~\cite{vaswani2017attention}, this class of neural language models follows an attention mechanism to learn the context of an overall input sequence. Our work focusses on a compact representation of the contextual embeddings by means of subspace embeddings, where we divide the contextual embedding in a way that restrains the generic context of the embedding.\\
\textbf{Language Models:} Generally, tokenizers are employed in language models to divide a sequence into tokens. Traditional tokenizers are crafted to split a sequence into contextual entities such as characters, morphs, and words. Such tokenizers can be implemented inherently, however, they are prone to out-of-vocabulary issues and require peculiar knowledge to divide into morphs. Existing work~\cite{sennrich2016neural} introduces that the tokenizers generate formed vocabularies via learning the input data. To overcome the aforementioned problem, the byte pair encoding~\cite{sennrich2016neural} is described to enfold all input cases~\cite{radford2019language}, and it iteratively consolidates tokens to a larger token. 
Canine~\cite{clark2022canine} solves the problem of the traditional tokenisation method, where the tokenizer is altered to character level in the absence of human knowledge. Similarly, Byt5~\cite{xue2022byt5} introduces token-free methods which encode input sequences without tokens and rely on contextual units. Our work presents the structure of embedding indifferent of tokenizers. It is different from token-free mechanism~\cite{xue2022byt5} in a way that our subspace embedding structure requires a tokenizer, but the PLMs do not endure from the available tokens. As the embeddings serve as an important actor in the language model, numerous approaches~\cite{lan2020albert} have been introduced to elucidate the embeddings, in terms of embedding compression, where an original embedding corresponds to the output tokens diversity. 
In our approach, we use a lookup operation to rebuild the embedding regardless of any further computations.
\vspace{-10pt}
\section{Subspace Embedding}
We present the formulation of our proposed approach and the methods of selecting subspace embeddings, including algorithmic descriptions of sharing the subspace embeddings. In Fig.~\ref{fig:subspacee}, we show that six subspace embedding vectors can be generated from eight embedding vectors. In this paper, we devise an embedding compression method through two devised algorithms. Firstly, we describe how to assign arbitrarily sporadic subspace embedding. Secondly, a cluster-based subspace embedding incorporates contextual information.
\vspace{-10pt}
\subsection{Problem Settings}\label{ssec:ps}
Initially, we divide the embedding vectors horizontally into subspace embedding (SE) vectors which are shared with certain different embedding vectors. In our proposed structure of embedding, the subspace embeddings recline to distinctly correlate. We devise two-fold steps to calibrate the subspace embedding similar to the original embeddings. a) Consider \(E_i\), \(E_j\) to be the original embedding vectors and their subspace embedding vectors are $\{v_{i}^{f}\}, \{v_{j}^{f}\}, \forall i,j \in \{1,2,...,D\}$, where $f\in \{1,2,...,F\}$ such that $\{v_{i}^{f}\} \neq \{v_{j}^{f}\}$ and $i \neq j$. This step is to verify that the partitioned embedding vectors are unique. b) Given a duet of tokens having identical contextual meanings, then their subspace embeddings allocate additional parts than a random duet. This hypothesis deals with the contextual mapping of subspace embedding. Based on the aforementioned steps, we then compute a function that maps the original embeddings into subspace embeddings. We imply a one-to-one correspondence function (or perfect pairing) $\mathcal{F}: \mathcal{P}\rightarrow\mathcal{Q}\times\ldots\times\mathcal{Q}$ for transforming the original embeddings to SE vectors where $\mathcal{P} \in \{1,2,...,D\} \subset \mathbb{N}$ represents a set of the embedding index, and $\mathcal{Q} = \{1,2,...,Q\}\subset\mathbb{N}$ describes a set of each SE vector index. This function can manifest for building subspace embedding vectors in many ways. Thus, the above mapping function can be generalised via the Cartesian product of functions as {\small$ \mathcal{F}(n) = (c_{1}\times c_{2}\times\ldots\times c_{f}) \underbrace{(n,\ldots,n)}_{f}$}, where the function $c_f: \mathcal{P}\rightarrow \mathcal{Q}$ delineates the \(f\)-th subspace embedding index. As the cardinality of \(f-\)ary Cartesian product is \(Q^{f}\), the embeddings can be inherent only \(D^{\frac{1}{f}}\) subspace embeddings. This construct can substantially reduce the number of embedding parameters with $\log$ scale. We consider the embedding dimension \(d\) to be allocated equally to subspace embedding $Q = d/f$. So, the number of embedding parameters that contain $D\times d$ embedding table is replaced with \(f\) distinct $Q\times (d/f)$ embedding table. Specifically, the embedding with \(d-\)dimensional vector for each token in the vocabulary is replaced by \(f\) varied subspace embedding vectors $\{v_{i}^{f}\}$, each of dimension $d/f$, where $v_i$ is drawn from the \(i-\)th fixed-size table of embedding vectors. Thus, the embedding representation can be formulated as $v_n = \oplus_{f=1,\ldots,F}v_{c_{f}(n)}$, where $v_n, v_{c_{f}}$ are the corresponding embedding vectors and $\oplus$ denotes the concatenation operation.
\vspace{-10pt}
\subsection{Arbitrarily Dispersed Subspace Embedding}
We establish $c_f$ to incorporate subspace embedding via a Cartesian product which can procreate up to \(Q^f\) embedding vectors. For the procreated embeddings to be unique, the number of each subspace embedding \(Q\) should be larger than \(D^{1/f}\). An algorithmic description of arbitrarily assigning subspace embedding in a sequential manner is reported in Algorithm~\ref{alg:subspace}. It continually uses the modulo operation  to procreate the entire embeddings. Based on the aforementioned assumption, we apply \(Q\)  = $\lceil$ \(D^{1/f}\) $\rceil$, where $\lceil . \rceil$ is a ceiling function, it is used to extract the most compact subspace embedding (the compressed form of an original embedding). This viewpoint is used to obtain the modulo operation through \(Q\)-base number. The transformation to \(Q\)-base number is the one-to-one correspondence function and we use each subspace embedding index as each digit of the base (or radix).
\vspace{-10pt}
\begin{algorithm}
\small
\caption{Assign Subspace Embedding Arbitrarily}\label{alg:subspace}
\begin{algorithmic}[1]
\Require \(D\) number of embeddings with dimension \(d\), and set of subspace embeddings \(F\)
\State \(Q\) $\leftarrow$ $\lceil$ \(D^{1/f}\) $\rceil$ \Comment{number of each subspace embedding}
\State Initialise \(f\)-th \(Q\) subspace embedding vectors $\{v_{q}^{f} \in\mathbb{R}^{\frac{d}{f}}$\}$_{q=1}^{Q}, \forall f\in\{1,\ldots,F\}$
\For {$n = 1,2,\ldots,D$}
    \For {f = 1,2,$\ldots,F$}
        \State $c_{f}(n) = (n/Q^{f-1})\mod Q^f$
    \EndFor
    \State $v_n = \oplus_{f=1}^{F}v_{c_{f}(n)}$
\EndFor
\Ensure The incorporated embedding vectors are $\{v_{n}\}_{n=1}^{D}$.
\end{algorithmic}
\end{algorithm}
\vspace{-20pt}
\subsection{Cluster-based Subspace Embedding}
This approach re-establish the subspace embedding based on contextual information from a pre-trained model. Recent advances~\cite{vaswani2017attention} in exploiting contextual information stem from the attention-based model, namely, Transformer. It learns the entire context of an input sequence where each token's embedding vector is mapped to the embedding space manifesting its context. In Word2Vec~\cite{mikolov2013distributed}, they show that two randomly mapped word vectors tend to have identical contexts. So, if each token has given its context, we can amplify the allotment heuristic via the contexts. In addition to tokens that have an identical meaning, we consider that the two tokens can be identified with smaller adjustments. Therefore, we can assign more subspace embedding to be shared, and the similarity of each duet of the tokens can be computed using a pre-trained model. The pre-trained model is employed to estimate the L2 distance among each embedding vector. Our conjecture is that all subspace embeddings are independently assigned arbitrarily, including, the tokens allocating more subspace embeddings that are anticipated to have less L2 distance. The technique of assigning subspace embedding to identical tokens follows the k-means clustering algorithm~\cite{arthur2006k}. Using k-means, we serve the embedding vectors as an instance of this clustering algorithm, and so the algorithm is altered iteratively to each subspace embedding vector. The iterative k-means purpose is to distinct the instances which are assigned in certain similar subspace embeddings. We describe our Algorithm~\ref{alg:cluster}, the mapping of subspace embedding can propitiate the second step(in Section~\ref{ssec:ps}) as the k-means algorithm is based on L2 norm.
\vspace{-10pt}
\begin{algorithm}
\small
\caption{Cluster-based Subspace Embedding}\label{alg:cluster}
\begin{algorithmic}[1]
\Require \(D\) number of embeddings, \(Q\) number of subspace embeddings, \(d\) dimension of embedding, and number of subspace embeddings set \(F\), the pre-trained embedding model $\mathcal{L}_P = \{p_{n}\}_{n=1}^{D}$
\State Initialise \(f\)-th \(Q\) subspace embedding vectors $\{v_{q}^{f} \in\mathbb{R}^{\frac{d}{f}}$\}$_{q=1}^{Q}, \forall f\in\{1,\ldots,F\}$
\State $c_f(n) \leftarrow 0, \forall f = 1,\ldots,F, n=1,\ldots,D$
\For {$f = 1,2,\ldots,F$}
    \State extract distinct tuples from $\{\mathcal{F}(n)\}_{n=1}^{D}$
    \For {distinct $\mathcal{F}(n^{*})$ in $\{\mathcal{F}(n)\}_{n=1}^{D}$}
        \If {$f\neq F$}
            \State $\{\mathcal{L}_P\}_{\mathcal{F}(n^{*})}\leftarrow \{p_{n}: \mathcal{F}(n) = \mathcal{F}(n^{*})\}_{n=1}^{D}$
            \State alter k-means algorithm to $\{\mathcal{L}_P\}_{\mathcal{F}(n^{*})}$
            \State the outcomes labelling to $c_f(n)$, where $\mathcal{F}(n) = \mathcal{F}(n^{*})$
        \Else
            \State $c_f(n)\leftarrow$ arbitrary number among \(Q\) candidates
        \EndIf
    \EndFor
\EndFor
\State Collect $v_n = \oplus_{f=1}^{F}v_{c_{f}(n)}, \forall n\in \mathcal{P}$
\Ensure The incorporated embedding vectors are $\{v_{n}\}_{n=1}^{D}$.
\end{algorithmic}
\end{algorithm}
\section{Experiments}
The experiments embark with substituting the original word embeddings of masked language modelling (MLM)~\cite{liu2019roberta,kenton2019bert,lan2020albert} with our subspace embedding to calibrate the impact of our proposed method. There exist some varied language models such as causal language and translation language modelling.
\\
\textbf{Dataset:} The language models are mainly trained with monolingual datasets and substantially fine-tuned on certain downstream tasks. Our work employs the multilingual dataset from Web Crawl~\cite{wenzek2020ccnet} in which we extract ten languages corpuses explicated in XNLI~\cite{conneau2018xnli}. In addition to it, we employ the monolingual datasets - Books~\cite{zhu2015aligning} and English Wikipedia corpuses\footnote{https://linguatools.org/tools/corpora/wikipedia-monolingual-corpora/} to conduct the evaluation of our proposed subspace embedding whether it supports large vocabularies.
\vspace{-20pt}
\subsection{Language Model Settings}
Our work employs the masked language modelling structure for the embedding network without using next-sentence prediction. The reason is due to the fact that token prediction networks such as MLM require language models decoder to identify token representation. Several language embeddings couple join the last output weights to the input embedding weights. Embedding coupling~\cite{chung2020rethinking} investigated that the decoder independent from the embeddings can be strengthened in terms of performance based on the decoder features. The coupling weights are alike the result of decoupling the decoder and the embedding. In our case, we trained the language models from coupling decoders. Furthermore, we substitute the embedding portion with the previous network into the subspace embedding model. There are still additional embeddings to apprehend external information, including token-type embeddings, and positional embeddings, however, we do not substitute these embeddings for the subspace embeddings. We alter the implementation of RoBERTa~\cite{liu2019roberta} and XLM-R~\cite{conneau2020unsupervised} models based on attention-based networks framework~\cite{wolf2020transformers}. Similar to other language models, we employ tokenizers in our approach. However, the tokenizer utilised in our model offers the advantage of easily incorporating new vocabularies through the combination of subspace embeddings. Consequently, our proposed approach is immune to the OOV problem. Our embedding network employs the hyperparameters from RoBERTa with the masking token probability of 0.15. Our base model comprises eight transformer encoder layers with 512-dimension embeddings in paucity to BERT-base. For multilingual cases, XLM-R is altered to eight layers as RoBERTa. The altered networks are indicated as $\text{RoBERTa}_{S}$ and $\text{XLM-R}_{S}$, where subscript \text{S} refers to subspace embedding. We present an arbitrarily assigned scenario of f-subspace embeddings reported in Table~\ref{tab:tab1}. As we show that the number of embedding parameters in the altered models is substantially reduced from $\text{RoBERTa}_{S}$, and $\text{XLM-R}_{S}$. Our tokenizer's configuration and training settings follow~\cite{liu2019roberta,conneau2020unsupervised}. 
\vspace{-10pt}
\begin{table}[htb]
\caption{Description of the altered neural language models.}\label{tab:tab1}
\vspace{-8pt}
\begin{adjustbox}{width=\linewidth}
\begin{tabular}{|c|c|c|c|c|}
\hline
NLMs &  Vocabulary Size & \# Embeddings & $\mid\theta\mid$ & $\mid\theta_{v}\mid$ \\
\hline
$\text{RoBERTa}_{S}$ &  50k & 50k & 51M & 25.7M \\
+2-SE &  50k & 225 & 26M & 115k\\
+3-SE & 50k & 37 & 26M & 18.9k\\
+8-SE & 50k & 4 & 26M & 2k\\ \hline
$\text{XLM-R}_{S}$ & 250k & 250k & 154M & 128M \\
+3-SE & 250k & 63 & 26M & 32k\\
\hline
\end{tabular}
\end{adjustbox}
\end{table}
\vspace{-18pt}
\subsection{Benchmarks}
The evaluations of altered neural language models are conducted through GLUE~\cite{wang2018glue} benchmark which comprises similarity, paraphrasing, single-sentence, and inference tasks. For multilingual language models, we employ the XNLI benchmark for evaluation, including, the fine-tuning of the pre-trained XLM-R in both manner, multi-language and cross-lingual tasks.
\\
\textbf{Results of Algorithm~\ref{alg:subspace}:} Our model $\text{RoBERTa}_{S}$ follow the base network based on~\cite{liu2019roberta}, where we substitute the input embedding component through arbitrarily dispersed \(f\)-subspace embedding constructs reported in Algorithm~\ref{alg:subspace}. We present our model (the base model and \(f\)-subspace embedding) results on the GLUE benchmark. It uses the number of SE as determined which is \(Q\)  = \(D^{1/f}\) and requires only 4 SE vectors provided \(f\) is 8 if \(D\) is 50,627. In Table~\ref{tab:tab_1}, the row below \(f\) represents the number of embeddings which are 50k for our model. The results on the GLUE benchmark reflect that \(f\)-SE models are having alike performance among each other. However, the results of SE models are relatively lower than the $\text{RoBERTa}_{S}$. This shows that the arbitrarily assigned dispersed subspace embeddings could not alleviate the distinctly entangled part of embedding. Also, we tend to not use certain special tokens, including, padding and separate tokens during assigning the subspace embeddings. It degrades the performance due to the aforementioned reasons. To address this problem, we devised a cluster-based allotment approach that employs contextual information from the pre-trained models.
\\
\textbf{Results of Algorithm~\ref{alg:cluster}:} We have shown above that our embedding compression method using arbitrarily dispersed subspace embedding successfully lightens the original embeddings. However, in certain cases, the performance degrades due to distinctly entangled parts of subspace embeddings. Though, the tokens situated in the latent space suffer to signify the context among them. To alleviate such context mismatching problem, we employ the pre-trained RoBERTa~\cite{liu2019roberta} model from ~\cite{wolf2020transformers}. We employ this pre-trained network with 768-d embedding vectors to apply our Algorithm~\ref{alg:cluster} for clustering. Our evaluation on GLUE benchmarks uses the number of subspace embedding \(Q\) with 50, 100, and 200 to assign 3-SE vectors. The results are reported in Table~\ref{tab:tab_2}. Based on the k-means clustering which uses instances rather than the cluster size, i.e. across varied k-clusters possess different cluster sizes. In our evaluation, we apply both naive k-means and uniformly allocated k-means as per Algorithm~\ref{alg:cluster}. The results found that the subspace embedding with uniform cluster size outperforms a small set of embedding. In particular, even with a 3-SE network after clustering has limited parameters than a 2-SE network, it is superior in terms of performance over each GLUE benchmark. Our proposed approach shows comparable performance with the original embedding model and is superior on SST-2 dataset. 
\\
\textbf{Results on Multilingual Dataset:} Multilingual language models are resource intensive, especially the training aspect as compared to the monolingual scenario. We use the XLM-R model based on the Unicoder~\cite{huang2019unicoder} to evaluate a cross-lingual transfer task. Our altered $\text{XLM-R}_{S}$ network with 250k and 63 number of embeddings for 3-subspace embedding, and 128 for clustered SE. On English dataset, their performances are 74\%, 72.6\%, and 72.9\% for $\text{XLM-R}_{S}$, 3-SE, and clustered SE. The results on XNLI are improved by 2\% on the cross-lingual transfer task, while the embeddings are compressed over 99.95\%.
\begin{table}[h]
    \centering
    \caption{Results of Arbitrarily Dispersed Subspace Embedding on GLUE. Shaded columns in blue colour is based on Algorithm~\ref{alg:subspace}.}\label{tab:tab_1}
   \vspace{-8pt}
    \begin{adjustbox}{width=\linewidth}
    \begin{NiceTabular}{|c|c|c|c|c|c|c|}\cline{2-7}
    \hline
    \multirow{3}{*}{\textbf{\diagbox{Dataset}{Model\\f}}} &
    $\text{RoBERTa}_{S}$~(Ours) & \Block[tikz={top color=blue!15}]{*-1}{}+2-SE & \Block[tikz={top color=blue!15}]{*-1}{}+3-SE & \Block[tikz={top color=blue!15}]{*-1}{}+4-SE & \Block[tikz={top color=blue!15}]{*-1}{}+6-SE & \Block[tikz={top color=blue!15}]{*-1}{}+8-SE \\
    \cline{2-7}
    & 1 & 2 & 3 & 4 & 6 & 8 \\
    & 50k & 225 & 37 & 15 & 7 & 4 \\
    \hline
    
    SST-2~\cite{socher2013recursive} & 89.8 & 88.4 & 88.0 & 88.1 & 87.2 & 88.0 \\
    Quora Questions\footnote{https://quoradata.quora.com/First-Quora-Dataset-Release-Question-Pairs} & 86.5 & 84.0 & 83.0 & 83.3 & 82.6 & 83.0 \\
    MNLI~\cite{williams2018broad} & 79.5 & 74.3 & 73.1 & 72.8 & 73.5 & 73.0 \\
    QNLI~\cite{rajpurkar2016squad} & 88.1 & 84.0 & 83.4 & 84.1 & 84.1 & 83.0 \\
    MRPC~\cite{dolan2005automatically} & 88.3 & 88.0 & 85.5 & 87.4 & 85.2 & 86.3 \\
    RTE~\cite{dagan2005pascal} & 72.8 & 66.9 & 67.8 & 70.0 & 67.4 & 67.8 \\
    STS-B~\cite{cer2017semeval} & 88.0 & 79.2 & 77.3 & 78.4 & 79.5 & 76.4 \\
    CoLA~\cite{warstadt2019neural} & 38.0 & 35.6 & 18.5 & 23.2 & 25.5 & 20.0 \\
    \hline
    \end{NiceTabular}
    \end{adjustbox}
\end{table}

\begin{table}[h!]
    \caption{Results of Cluster-based Subspace Embedding on GLUE. Shaded columns in red and yellow colour denotes the clustered SE using k-means, and uniform cluster size.}\label{tab:tab_2}
    \vspace{-8pt}
    \begin{adjustbox}{width=\linewidth}
    \begin{NiceTabular}{|c|c|c|c|c|c|c|c|}\cline{2-8}
    \hline
    \multirow{3}{*}{\textbf{\diagbox{Dataset}{Model\\f}}} &
    $\text{RoBERTa}_{S}$~(Ours) & \Block[tikz={top color=blue!15}]{*-1}{}+2-SE & \Block[tikz={top color=blue!15}]{*-1}{}+3-SE & \Block[tikz={top color=red!15}]{*-1}{}+3-SE & \Block[tikz={top color=red!15}]{*-1}{}+3-SE & \Block[tikz={top color=yellow!15}]{*-1}{}+3-SE & \Block[tikz={top color=yellow!15}]{*-1}{}+3-SE \\
    \cline{2-8}
    & 1 & 2 & 3 & 3 & 3 & 3 & 3 \\
    &  &  &  & Q=100 & Q=200 & Q=50 & Q=100 \\ \hline
    $\mid\theta_{v}\mid$ & 25.7M & 115k & 18.9k & 104k & 154k & 25.6k & 51.2k \\
    $\%\downarrow$ & - & 99.5 & 99.93 & 99.6 & 99.3 & 99.87 &99.8 
    \\
    \hline
    \hline
    SST-2~\cite{socher2013recursive} & 89.8 & 88.4 & 88.0 & 88.2 & 90.0 & 89.3 & 89.3 \\
    Quora Questions\footnote{https://quoradata.quora.com/First-Quora-Dataset-Release-Question-Pairs} & 86.5 & 84.0 & 83.0 & 84.7 & 85.6 & 84.5 & 84.6 \\
    MNLI~\cite{williams2018broad} & 79.5 & 74.3 & 73.1 & 75.9 & 77.5 & 75.8 & 77.2 \\
    QNLI~\cite{rajpurkar2016squad} & 88.1 & 84.0 & 83.4 & 85.1 & 85.5 & 83.5 & 85.8 \\
    MRPC~\cite{dolan2005automatically} & 88.3 & 88.0 & 85.5 & 87.3 & 88.6 & 87.7 & 87.3 \\
    RTE~\cite{dagan2005pascal} & 72.8 & 66.9 & 67.8 & 67.1 & 69.7 & 67.9 & 70.7 \\
    STS-B~\cite{cer2017semeval} & 88.0 & 79.2 & 77.3 & 81.6 & 84.5 & 80.1 & 84.8 \\
    CoLA~\cite{warstadt2019neural} & 38.0 & 35.6 & 18.5 & 37.5 & 34.9 & 33.6 & 36.7 \\
    \hline
    \end{NiceTabular}
    \end{adjustbox}
\end{table}

\vspace{-5pt}
\section{Conclusion}
This paper introduced a novel compact embedding structure to lighten the neural language models with its ability for training with far fewer parameters than the original embeddings. We devise two methods to assign shared subspace embedding to the embedding vector, of which, the first way is to allocate sequentially using the modulo operation (Algorithm~\ref{alg:subspace}). The second approach is to assign dispersed subspace embedding using the pre-trained language model that incorporates contextual information (Algorithm~\ref{alg:cluster}). The compressed subspace embedding significantly reduces the number of parameters by over 99\% due to incorporated embedding can be procreated exponentially via Cartesian product. Therefore, these lightweight embeddings (subspace embedding) perform better on GLUE and XNLI and is comparable to the base result. Our evaluation is conducted to substitute the embeddings in MLM using \(f\)-subspace embedding.


\bibliographystyle{ACM-Reference-Format}
\bibliography{sample-base}


\begin{thebibliography}{32}


\ifx \showCODEN    \undefined \def \showCODEN     #1{\unskip}     \fi
\ifx \showDOI      \undefined \def \showDOI       #1{#1}\fi
\ifx \showISBNx    \undefined \def \showISBNx     #1{\unskip}     \fi
\ifx \showISBNxiii \undefined \def \showISBNxiii  #1{\unskip}     \fi
\ifx \showISSN     \undefined \def \showISSN      #1{\unskip}     \fi
\ifx \showLCCN     \undefined \def \showLCCN      #1{\unskip}     \fi
\ifx \shownote     \undefined \def \shownote      #1{#1}          \fi
\ifx \showarticletitle \undefined \def \showarticletitle #1{#1}   \fi
\ifx \showURL      \undefined \def \showURL       {\relax}        \fi
\providecommand\bibfield[2]{#2}
\providecommand\bibinfo[2]{#2}
\providecommand\natexlab[1]{#1}
\providecommand\showeprint[2][]{arXiv:#2}

\bibitem[Arthur and Vassilvitskii(2006)]%
        {arthur2006k}
\bibfield{author}{\bibinfo{person}{David Arthur} {and} \bibinfo{person}{Sergei
  Vassilvitskii}.} \bibinfo{year}{2006}\natexlab{}.
\newblock \bibinfo{booktitle}{\emph{k-means++: The advantages of careful
  seeding}}.
\newblock \bibinfo{type}{{T}echnical {R}eport}.
\newblock


\bibitem[Bojanowski et~al\mbox{.}(2017)]%
        {bojanowski2017enriching}
\bibfield{author}{\bibinfo{person}{Piotr Bojanowski}, \bibinfo{person}{Edouard
  Grave}, \bibinfo{person}{Armand Joulin}, {and} \bibinfo{person}{Tomas
  Mikolov}.} \bibinfo{year}{2017}\natexlab{}.
\newblock \showarticletitle{Enriching word vectors with subword information}.
\newblock \bibinfo{journal}{\emph{Transactions of the association for
  computational linguistics}}  \bibinfo{volume}{5} (\bibinfo{year}{2017}),
  \bibinfo{pages}{135--146}.
\newblock


\bibitem[Cer et~al\mbox{.}(2017)]%
        {cer2017semeval}
\bibfield{author}{\bibinfo{person}{Daniel Cer}, \bibinfo{person}{Mona Diab},
  \bibinfo{person}{Eneko Agirre}, \bibinfo{person}{I{\~n}igo Lopez-Gazpio},
  {and} \bibinfo{person}{Lucia Specia}.} \bibinfo{year}{2017}\natexlab{}.
\newblock \showarticletitle{SemEval-2017 Task 1: Semantic Textual Similarity
  Multilingual and Crosslingual Focused Evaluation}. In
  \bibinfo{booktitle}{\emph{Proceedings of the 11th International Workshop on
  Semantic Evaluation (SemEval-2017)}}. \bibinfo{pages}{1--14}.
\newblock


\bibitem[Chung et~al\mbox{.}(2020)]%
        {chung2020rethinking}
\bibfield{author}{\bibinfo{person}{Hyung~Won Chung}, \bibinfo{person}{Thibault
  Fevry}, \bibinfo{person}{Henry Tsai}, \bibinfo{person}{Melvin Johnson}, {and}
  \bibinfo{person}{Sebastian Ruder}.} \bibinfo{year}{2020}\natexlab{}.
\newblock \showarticletitle{Rethinking Embedding Coupling in Pre-trained
  Language Models}. In \bibinfo{booktitle}{\emph{International Conference on
  Learning Representations}}.
\newblock


\bibitem[Clark et~al\mbox{.}(2022)]%
        {clark2022canine}
\bibfield{author}{\bibinfo{person}{Jonathan~H Clark}, \bibinfo{person}{Dan
  Garrette}, \bibinfo{person}{Iulia Turc}, {and} \bibinfo{person}{John
  Wieting}.} \bibinfo{year}{2022}\natexlab{}.
\newblock \showarticletitle{CANINE: Pre-training an Efficient Tokenization-Free
  Encoder for Language Representation}.
\newblock \bibinfo{journal}{\emph{Transactions of the Association for
  Computational Linguistics}}  \bibinfo{volume}{10} (\bibinfo{year}{2022}),
  \bibinfo{pages}{73--91}.
\newblock


\bibitem[Conneau et~al\mbox{.}(2020)]%
        {conneau2020unsupervised}
\bibfield{author}{\bibinfo{person}{Alexis Conneau}, \bibinfo{person}{Kartikay
  Khandelwal}, \bibinfo{person}{Naman Goyal}, \bibinfo{person}{Vishrav
  Chaudhary}, \bibinfo{person}{Guillaume Wenzek}, \bibinfo{person}{Francisco
  Guzm{\'a}n}, \bibinfo{person}{{\'E}douard Grave}, \bibinfo{person}{Myle Ott},
  \bibinfo{person}{Luke Zettlemoyer}, {and} \bibinfo{person}{Veselin
  Stoyanov}.} \bibinfo{year}{2020}\natexlab{}.
\newblock \showarticletitle{Unsupervised Cross-lingual Representation Learning
  at Scale}. In \bibinfo{booktitle}{\emph{Proceedings of the 58th Annual
  Meeting of the Association for Computational Linguistics}}.
  \bibinfo{pages}{8440--8451}.
\newblock


\bibitem[Conneau et~al\mbox{.}(2018)]%
        {conneau2018xnli}
\bibfield{author}{\bibinfo{person}{Alexis Conneau}, \bibinfo{person}{Ruty
  Rinott}, \bibinfo{person}{Guillaume Lample}, \bibinfo{person}{Adina
  Williams}, \bibinfo{person}{Samuel Bowman}, \bibinfo{person}{Holger Schwenk},
  {and} \bibinfo{person}{Veselin Stoyanov}.} \bibinfo{year}{2018}\natexlab{}.
\newblock \showarticletitle{XNLI: Evaluating Cross-lingual Sentence
  Representations}. In \bibinfo{booktitle}{\emph{Proceedings of the 2018
  Conference on Empirical Methods in Natural Language Processing}}.
  \bibinfo{pages}{2475--2485}.
\newblock


\bibitem[Dagan et~al\mbox{.}(2005)]%
        {dagan2005pascal}
\bibfield{author}{\bibinfo{person}{Ido Dagan}, \bibinfo{person}{Oren Glickman},
  {and} \bibinfo{person}{Bernardo Magnini}.} \bibinfo{year}{2005}\natexlab{}.
\newblock \showarticletitle{The PASCAL recognising textual entailment
  challenge}. In \bibinfo{booktitle}{\emph{Proceedings of the First
  international conference on Machine Learning Challenges: evaluating
  Predictive Uncertainty Visual Object Classification, and Recognizing Textual
  Entailment}}. \bibinfo{pages}{177--190}.
\newblock


\bibitem[Dolan and Brockett(2005)]%
        {dolan2005automatically}
\bibfield{author}{\bibinfo{person}{William~B Dolan} {and}
  \bibinfo{person}{Chris Brockett}.} \bibinfo{year}{2005}\natexlab{}.
\newblock \showarticletitle{Automatically Constructing a Corpus of Sentential
  Paraphrases}. In \bibinfo{booktitle}{\emph{Proceedings of the Third
  International Workshop on Paraphrasing (IWP2005)}}.
\newblock


\bibitem[Huang et~al\mbox{.}(2019)]%
        {huang2019unicoder}
\bibfield{author}{\bibinfo{person}{Haoyang Huang}, \bibinfo{person}{Yaobo
  Liang}, \bibinfo{person}{Nan Duan}, \bibinfo{person}{Ming Gong},
  \bibinfo{person}{Linjun Shou}, \bibinfo{person}{Daxin Jiang}, {and}
  \bibinfo{person}{Ming Zhou}.} \bibinfo{year}{2019}\natexlab{}.
\newblock \showarticletitle{Unicoder: A Universal Language Encoder by
  Pre-training with Multiple Cross-lingual Tasks}. In
  \bibinfo{booktitle}{\emph{Proceedings of the 2019 Conference on Empirical
  Methods in Natural Language Processing and the 9th International Joint
  Conference on Natural Language Processing (EMNLP-IJCNLP)}}.
  \bibinfo{pages}{2485--2494}.
\newblock


\bibitem[Kenton and Toutanova(2019)]%
        {kenton2019bert}
\bibfield{author}{\bibinfo{person}{Jacob Devlin Ming-Wei~Chang Kenton} {and}
  \bibinfo{person}{Lee~Kristina Toutanova}.} \bibinfo{year}{2019}\natexlab{}.
\newblock \showarticletitle{BERT: Pre-training of Deep Bidirectional
  Transformers for Language Understanding}. In
  \bibinfo{booktitle}{\emph{Proceedings of NAACL-HLT}}.
  \bibinfo{pages}{4171--4186}.
\newblock


\bibitem[Kudo and Richardson(2018)]%
        {kudo2018sentencepiece}
\bibfield{author}{\bibinfo{person}{Taku Kudo} {and} \bibinfo{person}{John
  Richardson}.} \bibinfo{year}{2018}\natexlab{}.
\newblock \showarticletitle{SentencePiece: A simple and language independent
  subword tokenizer and detokenizer for Neural Text Processing}. In
  \bibinfo{booktitle}{\emph{Proceedings of the 2018 Conference on Empirical
  Methods in Natural Language Processing: System Demonstrations}}.
  \bibinfo{pages}{66--71}.
\newblock


\bibitem[Lan et~al\mbox{.}(2020)]%
        {lan2020albert}
\bibfield{author}{\bibinfo{person}{Zhenzhong Lan}, \bibinfo{person}{Mingda
  Chen}, \bibinfo{person}{Sebastian Goodman}, \bibinfo{person}{Kevin Gimpel},
  \bibinfo{person}{Piyush Sharma}, {and} \bibinfo{person}{Radu Soricut}.}
  \bibinfo{year}{2020}\natexlab{}.
\newblock \showarticletitle{ALBERT: A Lite BERT for Self-supervised Learning of
  Language Representations}. In \bibinfo{booktitle}{\emph{International
  Conference on Learning Representations}}.
\newblock


\bibitem[Liu et~al\mbox{.}(2019)]%
        {liu2019roberta}
\bibfield{author}{\bibinfo{person}{Yinhan Liu}, \bibinfo{person}{Myle Ott},
  \bibinfo{person}{Naman Goyal}, \bibinfo{person}{Jingfei Du},
  \bibinfo{person}{Mandar Joshi}, \bibinfo{person}{Danqi Chen},
  \bibinfo{person}{Omer Levy}, \bibinfo{person}{Mike Lewis},
  \bibinfo{person}{Luke Zettlemoyer}, {and} \bibinfo{person}{Veselin
  Stoyanov}.} \bibinfo{year}{2019}\natexlab{}.
\newblock \showarticletitle{Roberta: A robustly optimized bert pretraining
  approach}.
\newblock \bibinfo{journal}{\emph{arXiv preprint arXiv:1907.11692}}
  (\bibinfo{year}{2019}).
\newblock


\bibitem[Mikolov et~al\mbox{.}(2013)]%
        {mikolov2013distributed}
\bibfield{author}{\bibinfo{person}{Tomas Mikolov}, \bibinfo{person}{Ilya
  Sutskever}, \bibinfo{person}{Kai Chen}, \bibinfo{person}{Greg~S Corrado},
  {and} \bibinfo{person}{Jeff Dean}.} \bibinfo{year}{2013}\natexlab{}.
\newblock \showarticletitle{Distributed representations of words and phrases
  and their compositionality}.
\newblock \bibinfo{journal}{\emph{Advances in neural information processing
  systems}}  \bibinfo{volume}{26} (\bibinfo{year}{2013}).
\newblock


\bibitem[Pennington et~al\mbox{.}(2014)]%
        {pennington2014glove}
\bibfield{author}{\bibinfo{person}{Jeffrey Pennington},
  \bibinfo{person}{Richard Socher}, {and} \bibinfo{person}{Christopher~D
  Manning}.} \bibinfo{year}{2014}\natexlab{}.
\newblock \showarticletitle{Glove: Global vectors for word representation}. In
  \bibinfo{booktitle}{\emph{Proceedings of the 2014 conference on empirical
  methods in natural language processing (EMNLP)}}.
  \bibinfo{pages}{1532--1543}.
\newblock


\bibitem[Peters et~al\mbox{.}(2018)]%
        {peters-etal-2018-deep}
\bibfield{author}{\bibinfo{person}{Matthew~E. Peters}, \bibinfo{person}{Mark
  Neumann}, \bibinfo{person}{Mohit Iyyer}, \bibinfo{person}{Matt Gardner},
  \bibinfo{person}{Christopher Clark}, \bibinfo{person}{Kenton Lee}, {and}
  \bibinfo{person}{Luke Zettlemoyer}.} \bibinfo{year}{2018}\natexlab{}.
\newblock \showarticletitle{Deep Contextualized Word Representations}. In
  \bibinfo{booktitle}{\emph{Proceedings of the 2018 Conference of the North
  {A}merican Chapter of the Association for Computational Linguistics: Human
  Language Technologies, Volume 1 (Long Papers)}}.
  \bibinfo{publisher}{Association for Computational Linguistics},
  \bibinfo{address}{New Orleans, Louisiana}, \bibinfo{pages}{2227--2237}.
\newblock
\urldef\tempurl%
\url{https://doi.org/10.18653/v1/N18-1202}
\showDOI{\tempurl}


\bibitem[Radford et~al\mbox{.}(2019)]%
        {radford2019language}
\bibfield{author}{\bibinfo{person}{Alec Radford}, \bibinfo{person}{Jeffrey Wu},
  \bibinfo{person}{Rewon Child}, \bibinfo{person}{David Luan},
  \bibinfo{person}{Dario Amodei}, \bibinfo{person}{Ilya Sutskever},
  {et~al\mbox{.}}} \bibinfo{year}{2019}\natexlab{}.
\newblock \showarticletitle{Language models are unsupervised multitask
  learners}.
\newblock \bibinfo{journal}{\emph{OpenAI blog}} \bibinfo{volume}{1},
  \bibinfo{number}{8} (\bibinfo{year}{2019}), \bibinfo{pages}{9}.
\newblock


\bibitem[Rajpurkar et~al\mbox{.}(2016)]%
        {rajpurkar2016squad}
\bibfield{author}{\bibinfo{person}{Pranav Rajpurkar}, \bibinfo{person}{Jian
  Zhang}, \bibinfo{person}{Konstantin Lopyrev}, {and} \bibinfo{person}{Percy
  Liang}.} \bibinfo{year}{2016}\natexlab{}.
\newblock \showarticletitle{SQuAD: 100,000+ Questions for Machine Comprehension
  of Text}. In \bibinfo{booktitle}{\emph{Proceedings of the 2016 Conference on
  Empirical Methods in Natural Language Processing}}.
  \bibinfo{pages}{2383--2392}.
\newblock


\bibitem[Sennrich et~al\mbox{.}(2016)]%
        {sennrich2016neural}
\bibfield{author}{\bibinfo{person}{Rico Sennrich}, \bibinfo{person}{Barry
  Haddow}, {and} \bibinfo{person}{Alexandra Birch}.}
  \bibinfo{year}{2016}\natexlab{}.
\newblock \showarticletitle{Neural Machine Translation of Rare Words with
  Subword Units}. In \bibinfo{booktitle}{\emph{54th Annual Meeting of the
  Association for Computational Linguistics}}. Association for Computational
  Linguistics (ACL), \bibinfo{pages}{1715--1725}.
\newblock


\bibitem[Socher et~al\mbox{.}(2013)]%
        {socher2013recursive}
\bibfield{author}{\bibinfo{person}{Richard Socher}, \bibinfo{person}{Alex
  Perelygin}, \bibinfo{person}{Jean Wu}, \bibinfo{person}{Jason Chuang},
  \bibinfo{person}{Christopher~D Manning}, \bibinfo{person}{Andrew~Y Ng}, {and}
  \bibinfo{person}{Christopher Potts}.} \bibinfo{year}{2013}\natexlab{}.
\newblock \showarticletitle{Recursive deep models for semantic compositionality
  over a sentiment treebank}. In \bibinfo{booktitle}{\emph{Proceedings of the
  2013 conference on empirical methods in natural language processing}}.
  \bibinfo{pages}{1631--1642}.
\newblock


\bibitem[Sohler and Woodruff(2011)]%
        {sohler2011subspace}
\bibfield{author}{\bibinfo{person}{Christian Sohler} {and}
  \bibinfo{person}{David~P Woodruff}.} \bibinfo{year}{2011}\natexlab{}.
\newblock \showarticletitle{Subspace embeddings for the l1-norm with
  applications}. In \bibinfo{booktitle}{\emph{Proceedings of the forty-third
  annual ACM symposium on Theory of computing}}. \bibinfo{pages}{755--764}.
\newblock


\bibitem[Tay et~al\mbox{.}(2021)]%
        {tay2021charformer}
\bibfield{author}{\bibinfo{person}{Yi Tay}, \bibinfo{person}{Vinh~Q Tran},
  \bibinfo{person}{Sebastian Ruder}, \bibinfo{person}{Jai Gupta},
  \bibinfo{person}{Hyung~Won Chung}, \bibinfo{person}{Dara Bahri},
  \bibinfo{person}{Zhen Qin}, \bibinfo{person}{Simon Baumgartner},
  \bibinfo{person}{Cong Yu}, {and} \bibinfo{person}{Donald Metzler}.}
  \bibinfo{year}{2021}\natexlab{}.
\newblock \showarticletitle{Charformer: Fast Character Transformers via
  Gradient-based Subword Tokenization}. In
  \bibinfo{booktitle}{\emph{International Conference on Learning
  Representations}}.
\newblock


\bibitem[Vaswani et~al\mbox{.}(2017)]%
        {vaswani2017attention}
\bibfield{author}{\bibinfo{person}{Ashish Vaswani}, \bibinfo{person}{Noam
  Shazeer}, \bibinfo{person}{Niki Parmar}, \bibinfo{person}{Jakob Uszkoreit},
  \bibinfo{person}{Llion Jones}, \bibinfo{person}{Aidan~N Gomez},
  \bibinfo{person}{{\L}ukasz Kaiser}, {and} \bibinfo{person}{Illia
  Polosukhin}.} \bibinfo{year}{2017}\natexlab{}.
\newblock \showarticletitle{Attention is all you need}.
\newblock \bibinfo{journal}{\emph{Advances in neural information processing
  systems}}  \bibinfo{volume}{30} (\bibinfo{year}{2017}).
\newblock


\bibitem[Wang et~al\mbox{.}(2018)]%
        {wang2018glue}
\bibfield{author}{\bibinfo{person}{Alex Wang}, \bibinfo{person}{Amanpreet
  Singh}, \bibinfo{person}{Julian Michael}, \bibinfo{person}{Felix Hill},
  \bibinfo{person}{Omer Levy}, {and} \bibinfo{person}{Samuel Bowman}.}
  \bibinfo{year}{2018}\natexlab{}.
\newblock \showarticletitle{GLUE: A Multi-Task Benchmark and Analysis Platform
  for Natural Language Understanding}. In \bibinfo{booktitle}{\emph{Proceedings
  of the 2018 EMNLP Workshop BlackboxNLP: Analyzing and Interpreting Neural
  Networks for NLP}}. \bibinfo{pages}{353--355}.
\newblock


\bibitem[Warstadt et~al\mbox{.}(2019)]%
        {warstadt2019neural}
\bibfield{author}{\bibinfo{person}{Alex Warstadt}, \bibinfo{person}{Amanpreet
  Singh}, {and} \bibinfo{person}{Samuel~R Bowman}.}
  \bibinfo{year}{2019}\natexlab{}.
\newblock \showarticletitle{Neural network acceptability judgments}.
\newblock \bibinfo{journal}{\emph{Transactions of the Association for
  Computational Linguistics}}  \bibinfo{volume}{7} (\bibinfo{year}{2019}),
  \bibinfo{pages}{625--641}.
\newblock


\bibitem[Wenzek et~al\mbox{.}(2020)]%
        {wenzek2020ccnet}
\bibfield{author}{\bibinfo{person}{Guillaume Wenzek},
  \bibinfo{person}{Marie-Anne Lachaux}, \bibinfo{person}{Alexis Conneau},
  \bibinfo{person}{Vishrav Chaudhary}, \bibinfo{person}{Francisco Guzm{\'a}n},
  \bibinfo{person}{Armand Joulin}, {and} \bibinfo{person}{{\'E}douard Grave}.}
  \bibinfo{year}{2020}\natexlab{}.
\newblock \showarticletitle{CCNet: Extracting High Quality Monolingual Datasets
  from Web Crawl Data}. In \bibinfo{booktitle}{\emph{Proceedings of the 12th
  Language Resources and Evaluation Conference}}. \bibinfo{pages}{4003--4012}.
\newblock


\bibitem[Williams et~al\mbox{.}(2018)]%
        {williams2018broad}
\bibfield{author}{\bibinfo{person}{Adina Williams}, \bibinfo{person}{Nikita
  Nangia}, {and} \bibinfo{person}{Samuel Bowman}.}
  \bibinfo{year}{2018}\natexlab{}.
\newblock \showarticletitle{A Broad-Coverage Challenge Corpus for Sentence
  Understanding through Inference}. In \bibinfo{booktitle}{\emph{Proceedings of
  the 2018 Conference of the North American Chapter of the Association for
  Computational Linguistics: Human Language Technologies, Volume 1 (Long
  Papers)}}. \bibinfo{pages}{1112--1122}.
\newblock


\bibitem[Wolf et~al\mbox{.}(2020)]%
        {wolf2020transformers}
\bibfield{author}{\bibinfo{person}{Thomas Wolf}, \bibinfo{person}{Lysandre
  Debut}, \bibinfo{person}{Victor Sanh}, \bibinfo{person}{Julien Chaumond},
  \bibinfo{person}{Clement Delangue}, \bibinfo{person}{Anthony Moi},
  \bibinfo{person}{Pierric Cistac}, \bibinfo{person}{Tim Rault},
  \bibinfo{person}{R{\'e}mi Louf}, \bibinfo{person}{Morgan Funtowicz},
  {et~al\mbox{.}}} \bibinfo{year}{2020}\natexlab{}.
\newblock \showarticletitle{Transformers: State-of-the-art natural language
  processing}. In \bibinfo{booktitle}{\emph{Proceedings of the 2020 conference
  on empirical methods in natural language processing: system demonstrations}}.
  \bibinfo{pages}{38--45}.
\newblock


\bibitem[Wortsman et~al\mbox{.}(2021)]%
        {wortsman2021learning}
\bibfield{author}{\bibinfo{person}{Mitchell Wortsman},
  \bibinfo{person}{Maxwell~C Horton}, \bibinfo{person}{Carlos Guestrin},
  \bibinfo{person}{Ali Farhadi}, {and} \bibinfo{person}{Mohammad Rastegari}.}
  \bibinfo{year}{2021}\natexlab{}.
\newblock \showarticletitle{Learning neural network subspaces}. In
  \bibinfo{booktitle}{\emph{International Conference on Machine Learning}}.
  PMLR, \bibinfo{pages}{11217--11227}.
\newblock


\bibitem[Xue et~al\mbox{.}(2022)]%
        {xue2022byt5}
\bibfield{author}{\bibinfo{person}{Linting Xue}, \bibinfo{person}{Aditya
  Barua}, \bibinfo{person}{Noah Constant}, \bibinfo{person}{Rami Al-Rfou},
  \bibinfo{person}{Sharan Narang}, \bibinfo{person}{Mihir Kale},
  \bibinfo{person}{Adam Roberts}, {and} \bibinfo{person}{Colin Raffel}.}
  \bibinfo{year}{2022}\natexlab{}.
\newblock \showarticletitle{Byt5: Towards a token-free future with pre-trained
  byte-to-byte models}.
\newblock \bibinfo{journal}{\emph{Transactions of the Association for
  Computational Linguistics}}  \bibinfo{volume}{10} (\bibinfo{year}{2022}),
  \bibinfo{pages}{291--306}.
\newblock


\bibitem[Zhu et~al\mbox{.}(2015)]%
        {zhu2015aligning}
\bibfield{author}{\bibinfo{person}{Yukun Zhu}, \bibinfo{person}{Ryan Kiros},
  \bibinfo{person}{Rich Zemel}, \bibinfo{person}{Ruslan Salakhutdinov},
  \bibinfo{person}{Raquel Urtasun}, \bibinfo{person}{Antonio Torralba}, {and}
  \bibinfo{person}{Sanja Fidler}.} \bibinfo{year}{2015}\natexlab{}.
\newblock \showarticletitle{Aligning books and movies: Towards story-like
  visual explanations by watching movies and reading books}. In
  \bibinfo{booktitle}{\emph{Proceedings of the IEEE international conference on
  computer vision}}. \bibinfo{pages}{19--27}.
\newblock


\end{thebibliography}

\appendix

\end{document}